\documentclass[10pt, conference, letterpaper]{IEEEtran}

\usepackage{cite}
\usepackage{amsmath,amssymb,amsfonts}
\usepackage{algorithmic}
\usepackage{graphicx}
\usepackage{booktabs}
\usepackage{subcaption}            % 서브그림용 (modern). 캡션은 IEEE 스타일 유지됨.
\usepackage{textcomp}
\usepackage{xcolor}
\usepackage[hidelinks]{hyperref}   % 제출 전 PDF 메타데이터 제거 필수.
\graphicspath{{figures/}}          % 그림 파일은 figures/ 폴더에 두기.
\usepackage{comment}
\usepackage{adjustbox}
\usepackage{pifont}
\usepackage{bm}
\usepackage{enumitem}
\usepackage{multirow}
\usepackage{tabularx}

\newlength{\figninH}
\usepackage{siunitx}
\usepackage{tikz}
\newcommand*\circled[1]{\tikz[baseline=(char.base)]{
            \node[shape=circle,draw,inner sep=0.8pt] (char) {#1};}}

\newcommand{\system}{MPT}

\providecommand{\cmark}{\ding{51}}
\providecommand{\xmark}{\ding{55}}

\begin{document}
\typeout{>>> FIG9 HEIGHT = \the\figninH}

%---------------- Title (식별 가능한 프로젝트명 금지) ----------------
\title{\system{}: Missing Prototype Tracking via Barycentric Reconstruction in Vehicular Federated Learning}
%\title{Lightweight Barycentric Prototype Reconstruction in Vehicular Federated Learning}

%---------------- 저자: 리뷰본은 익명 (double-blind) ----------------
%\author{%
%  \IEEEauthorblockN{Anonymous Author(s)}
%  \IEEEauthorblockA{Submission for double-blind review\\
%  Author names and affiliations omitted}%
%}
% camera-ready 교체 예:
% \author{\IEEEauthorblockN{Name1, Name2}
%   \IEEEauthorblockA{Dept., Univ., City, Country, email}}

\newif\ifanon
\anonfalse   % arXiv 업로드 = \anonfalse   |   EDAS 더블블라인드 = \anontrue

\ifanon
  \author{%
    \IEEEauthorblockN{Anonymous Author(s)}
    \IEEEauthorblockA{Submission for double-blind review}%
  }
\else
  \author{%
    \IEEEauthorblockN{%
      Hanju Jang\IEEEauthorrefmark{1},
      Gyeongmin Han\IEEEauthorrefmark{1},
      Sungmin Lee\IEEEauthorrefmark{1},
      Kichang Lee\IEEEauthorrefmark{1},
      Chunghan Lee\IEEEauthorrefmark{2},
      JeongGil Ko\IEEEauthorrefmark{1}%
    }
    \IEEEauthorblockA{\IEEEauthorrefmark{1}School of Integrated Technology, Yonsei University, Republic of Korea}
    \IEEEauthorblockA{\IEEEauthorrefmark{2}Toyota Motor Corporation, Japan}
    \IEEEauthorblockA{\{unasemana, 
sd061123, kichang.lee, 
i.am.sungmin.lee. jeonggil.ko 
 \}@yonsei.ac.kr, 
lch@toyota-tokyo.tech}%
  }
  \hypersetup{%
    pdftitle={MPT: Missing Prototype Tracking via Barycentric Reconstruction in Vehicular Federated Learning},
    pdfauthor={Hanju Jang, Gyeongmin Han, Sungmin Lee, Kichang Lee, Chunghan Lee, JeongGil Ko}%
  }
\fi

\maketitle
%---------------- 섹션 블록 (순서 = main에서 제어) ----------------
%================================================================
% abstract.tex — Abstract + Keywords
% (main.tex의 \maketitle 다음에 input되어야 함)
%================================================================
\begin{abstract}
% 150-250 단어, 한 단락. 수식/인용 지양.
% (1) setting 한 줄 -> (2) 기존 가정이 깨지는 지점 ->
% (3) 무엇을 보였는가(측정된 결과만) -> (4) 함의 한 줄.

Cross-vehicle federated learning enables vehicles to collaboratively improve perception models while keeping locally collected driving data private. However, vehicle participation is transient, and a vehicle may depart before training converges while permanently taking its local data. When this departing vehicle holds most samples of a target class, the class becomes rare in the remaining FL network, and its recognition can silently degrade as the shared backbone continues to evolve. Recovering the class is difficult since the few remaining samples provide a noisy prototype estimate, while FL privacy constraints prevent centralized access to raw data or per-sample features. This paper presents \system{}, a cross-vehicle FL framework that maintains rare-class recognition by reconstructing its prototype at every round from privacy-preserving class-level statistics. \system{} combines a barycentric decomposition that tracks drift shared with remaining-class prototypes, a covariance-based residual prediction that estimates out-of-span drift, and an adaptive calibration that weighs the remaining rare-class samples according to their reliability. We evaluate \system{} on three vehicle classification tasks and four backbones against representative calibration and drift-compensation baselines. \system{} outperforms all baselines in rare class F1, reaching 0.516 on the nuImages dataset with only 1\% of rare-class samples remaining, without raw data, per-sample features, or retraining.

\end{abstract}

      % abstract + keywords (반드시 \maketitle 다음)
%================================================================
% intro.tex — Introduction (+ Contributions)
%================================================================
\section{Introduction}
\label{sec:intro}
Modern vehicles increasingly rely on onboard perception models for tasks such as object classification~\cite{chellapandi2023federated,chellapandi2023survey,park2023attfl}. As driving conditions evolve and vehicles encounter previously unseen environments, continuously updating these models using data collected across vehicles becomes a promising approach. However, sharing the raw data among vehicles incurs substantial communication costs and raises privacy  concerns~\cite{chellapandi2023federated, lee2024tazza}.
Federated learning (FL) offers a promising alternative by exchanging model parameters and compact auxiliary statistics while keeping data on each vehicle~\cite{zhai2024fedrav,chen2023data}.

\begin{figure}[t]
  \centering
  \includegraphics[width=0.95\columnwidth]{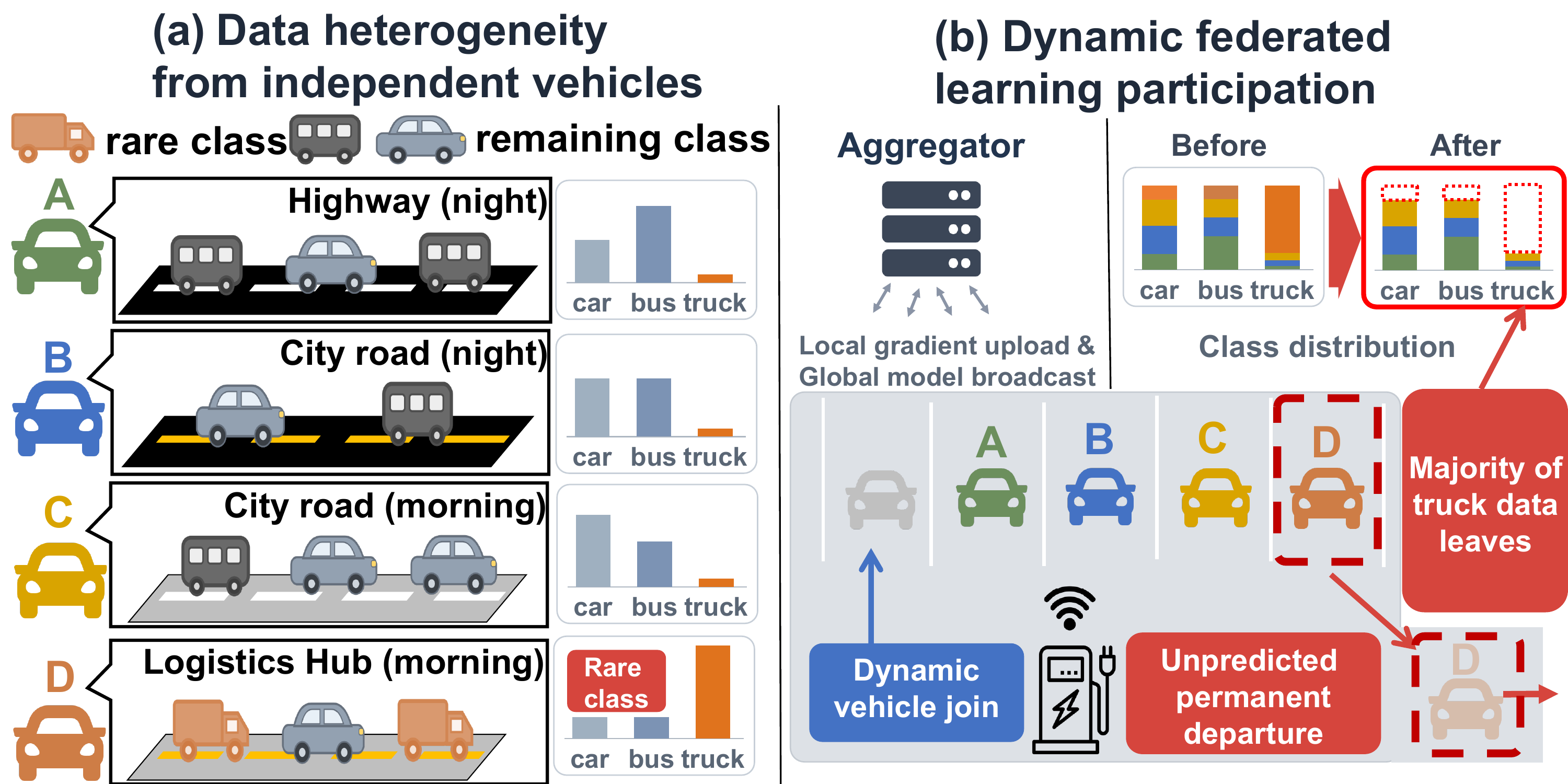}
  \caption{Target vehicular FL scenario overview.
  }
  \vspace{-3ex}
  \label{fig:scope}
\end{figure}

Figure~\ref{fig:scope} illustrates a representative \textit{cross-vehicle} FL scenario and its characteristics. As Figure~\ref{fig:scope} (a) shows, vehicles traveling along different routes encounter different objects, causing their local data distributions to substantially vary, and a single vehicle can dominate most samples of a specific class. At a charging facility, the vehicles use computing infrastructure and stable connectivity to form a spontaneous FL cluster (c.f., Fig.~\ref{fig:scope} (b))~\cite{wang2018edge,chellapandi2023federated}. However, here, a vehicle may leave at any training round according to its charging schedule, permanently taking its local data. When this \textit{departing vehicle} holds most samples of a target class, only a few samples remain on the other vehicles, causing it to be a \textit{rare class}.

Vehicle departure introduces challenges both before and after its occurrence. Before departure, since a vehicle may leave prior to FL convergence, the system must efficiently exploit its data within the limited participation window. However, achieving such rapid convergence is challenging, especially when most samples of a class dominated by a single vehicle. Under this skewed distribution, standard FL with a learnable classifier head (e.g., softmax) can favor frequently observed classes even when the backbone feature extractor (e.g., a CNN) has learned useful representations~\cite{luo2021no}. As a solution, \textit{prototype}-based methods mitigate classifier-head bias by leveraging relatively well-preserved feature spaces learned by the backbone~\cite{rebuffi2017icarl}. These methods represent each class using a prototype, typically computed as the mean feature representation of its samples. Prototypes are then used either to recalibrate a biased classifier head~\cite{luo2021no}, or to replace it with nearest-prototype classification, which assigns each input to the class with the closest prototype~\cite{rebuffi2017icarl}.

While prototype-based inference improves pre-departure performance, a new challenge emerges after the vehicle departure. As the shared backbone continuously evolves via updates from remaining vehicles, the stored rare-class prototype becomes increasingly misaligned with the current (updated) feature space. Correcting this drift is difficult given that the few remaining rare-class samples provide only a noisy estimate of an updated prototype, while continued FL training without sufficient target class samples further degrades its representation. Existing drift-compensation methods are also ill-suited to FL. Schemes such as Semantic Drift Compensation (SDC)~\cite{yu2020semantic} rely on reliable class-level references, which become unstable when rare-class samples are scarce and distributed across vehicles. On the other hand, Learnable Drift Compensation (LDC)~\cite{gomez2024exemplar} requires sample-level features extracted by both old and updated backbones, which cannot be centrally collected given FL data privacy goals~\cite{li2024fcs,rypesc2024task}. Thus, FL requires drift-compensation to update the rare-class prototype from limited remaining data while preserving data locality and privacy.

Motivated by these challenges, we propose \system{}, a prototype tracking framework for cross-vehicle FL that preserves a rare class concentrated on a vehicle that departs during training. \system{} uses a prototype-based Nearest-Class-Mean (NCM) classifier~\cite{mensink2013distance}, which provides a stable readout under highly skewed class distributions, and continuously updates the departed class's prototype using class-level statistics from remaining vehicles. 
The key idea is to decompose the rare-class prototype into two parts. The first is an \emph{in-span component}, expressed by its barycentric relationship to the remaining-class prototypes. The second is an \emph{out-of-span component}, captured by a residual that cannot be spanned by remaining classes. After each FL round, remaining vehicles re-embed local data using the updated backbone and report class-level prototypes and covariances. Upon vehicle departure, \system{} re-evaluates the in-span component using updated remaining-class prototypes and corrects the residual using covariance changes and any available rare-class samples on remaining vehicles. 
In this way, \system{} reconstructs the rare-class prototype throughout FL training using only privacy-preserving class-level statistics.

Specifically, this paper makes the following contributions:
 \begin{itemize}[leftmargin=*]
    \item We characterize cross-vehicle FL with heterogeneous data distributions and dynamic vehicle participation as requiring rapid convergence before departure and rare-class prototype tracking afterward, since a vehicle holding most samples of a class may depart while continued backbone updates cause representation drift.

    \item We propose \system{}, a prototype-based cross-vehicle FL framework for rapid and stable learning before departure and reconstructs the rare-class prototype afterward by tracking shared drift through its departure-time barycentric relationship, adapting the out-of-span residual from covariance changes, and calibrating it with aggregate statistics from remaining rare-class samples.

    \item We evaluate \system{} across three vehicle-classification tasks, four backbones, against five baseline architectures. Even with only 1\% of rare-class samples remaining after departure, \system{} achieves 51.6\% rare-class F1 on the nuImages dataset, outperforming the strongest baseline by 8.1\%p and exceeds the 50.5\% oracle. We show that tracking the barycentric relationship and adapting the out-of-span residual jointly reduce prototype error after departure.

\end{itemize}

\begin{comment}
\end{comment}

    % \textbf{Problem characterization and observability analysis.}
    % We identify the class-drop scenario in vehicular FL, where the permanent departure of the sole owner of a rare, confusable class induces a previously unrecognized failure mode. We show that the classifier collapses while the underlying feature representation remains largely preserved, and analyze the observability limits of prototype evolution after the drop.
    
    % \textbf{Prototype tracking framework.}
    % We propose \system{}, a communication-efficient prototype tracking framework that stores a compact one-time summary of a departed class and continuously re-aligns it using only aggregated class statistics already exchanged during federated learning, requiring no raw-data sharing, no per-sample features, and no learned drift models.
    
    % \textbf{Experimental evaluation.}
    % We evaluate \system{} on cropped nuScenes~\cite{caesar2020nuscenes} (nuImages) and Car1000~\cite{hu2025car} across multiple backbone architectures, monopoly levels, and departure scenarios. \system{} closely tracks the oracle prototype, matches or exceeds exemplar- and map-based alternatives, while not requiring raw data at the server.

%The rest this paper is organized as follows. Section~\ref{sec:related} reviews related work, Section~\ref{sec:prelim} presents the preliminary study, Section~\ref{sec:scheme} describes \system{}, Section~\ref{sec:eval} evaluates its performance, Section~\ref{sec:discussion} discusses its applicability and limitations, and Section~\ref{sec:conclusion} concludes the paper.  % Introduction + Contributions
\section{Related Work}
\label{sec:related}

% \sm{\cite{gu2021fast}?? unstable connectivity rejoin}
\noindent\textbf{Cross-vehicle federated learning.}
Federated learning (FL) enables multiple clients to collaboratively train a shared model without centralizing raw data~\cite{mcmahan2017communication}. In cross-vehicle FL, vehicles continuously collect driving data and opportunistically participate in federated training when parked and connected to the FL infrastructure. Existing work has demonstrated the practicality of this paradigm for collaborative perception and edge intelligence~\cite{zhai2024fedrav,chen2023data}. Unlike conventional cross-silo FL, vehicle participation is inherently transient. Thus, when a vehicle departs from the physical environment, its locally collected data permanently disappears from the federation, while the model continues to evolve; creating a unique knowledge preservation problem for the now \textit{rare class} data that becomes increasingly severe as FL training rounds progress.

\vspace{0.5ex}
\noindent\textbf{Classifier-head forgetting in federated learning.}
We note that such knowledge loss concentrates in the classifier head rather than in the feature extractor~\cite{DBLP:conf/nips/YosinskiCBL14,kang2019decoupling,DBLP:conf/cvpr/AlshammariWRK22,DBLP:conf/cvpr/ZhouCWC20,luo2021no,zhang2022federated}, since shared representations are updated by remaining vehicles while the departed data's class decision boundary receives only negative signals and silently erodes~\cite{gomez2024exemplar,wu2025demystifying}. Training-time interventions such as FedAvg~\cite{mcmahan2017communication} and FedLC~\cite{zhang2022federated} improve representation learning under non-IID participation, yet cannot create positive supervision for an already departed class. Post-hoc solutions such as CCVR~\cite{luo2021no} rebalance the classifier via calibration, personalization, exemplar replay, or retraining~\cite{DBLP:conf/kdd/LiZ21,zhang2022federated,DBLP:conf/iclr/OhKY22,DBLP:journals/corr/abs-2107-00778,DBLP:journals/corr/abs-1912-00818,DBLP:conf/icml/CollinsHMS21}, but require continued access to class-specific samples, gradients, or reliable statistics. As an alternative approach, prototype-based classifiers such as iCaRL~\cite{rebuffi2017icarl} directly represent each class in the feature space~\cite{mensink2013distance,rebuffi2017icarl,DBLP:conf/wacv/PetitPSPD23,DBLP:conf/nips/GoswamiL0W23}. However, when a departed class prototype is constructed from only a few remaining samples of the target/rare class, it gradually drifts from the live representation.

\vspace{0.5ex}
\noindent\textbf{Prototype drift compensation.}
A large body of work has sought to track this drift when using prototype-based classifiers. Some recompute prototypes from stored exemplars~\cite{rebuffi2017icarl} or freeze the backbone~\cite{DBLP:conf/wacv/PetitPSPD23,DBLP:conf/nips/GoswamiL0W23}, at the cost of storage and plasticity. Among compensation-based techniques, SDC transfers a weighted average of embedding displacements measured on available samples to the stored prototypes~\cite{yu2020semantic}, and LDC learns a mapping from the old feature space to the new one~\cite{gomez2024exemplar,li2024fcs,rypesc2024task}. These approaches, however, depend on sample-level features observed via both previous and updated backbone, which conflicts with FL's data-privacy constraints, where raw data stays with client~\cite{chellapandi2023federated,mcmahan2017communication,zhou2022ppa,zhu2019deep,geiping2020inverting}.

% 릴웍 재배치
% 1. cross-vehicle FL
% [SM: [5]?? 불안정한 연결성 재참여] 연합 학습(FL)은 원시 데이터를 중앙에 집중시키지 않고도 여러 클라이언트가 협력하여 공유 모델을 학습할 수 있게 한다 [6]. 차량 간 연합 학습(CVFL)에서 차량들은 주행 중 지속적으로 데이터를 수집하고, 주차되어 FL 인프라에 연결되었을 때 기회적으로(opportunistically) 공유 모델 학습에 참여한다. 기존 연구들은 협력적 인지(collaborative perception)와 엣지 인텔리전스 분야에서 이러한 패러다임의 실용성을 입증해 왔다 [3], [4]. 기존의 cross-silo FL과 달리, 차량의 참여는 본질적으로 일시적(transient)이다. 다시 말해, 차량이 물리적 환경에서 이탈하면 그 차량이 로컬에서 수집한 데이터는 연합에서 영구히 사라지는 반면, 모델은 계속해서 진화한다. 이는 학습이 진행될수록 점점 더 심각해지는 고유한 지식 보존(knowledge preservation) 문제를 야기한다.

% 2. (Head+Prototype based) classifier
% 이러한 지식 소실은 특징 추출기가 아니라 분류기 헤드에 집중된다 [7]–[12]. 클래스 간에 공유되는 특징 표현은 생존 차량들의 학습만으로도 대체로 유지되는 반면, 이탈 클래스의 결정 경계는 이탈 이후 오직 음(negative)의 신호만을 공급받아 조용히 침식되기 때문이다 [13], [14]. 기존 해법들은 보정, 개인화, exemplar 리플레이, 또는 사후 재학습을 통해 분류기를 재균형화하지만 [12], [15]–[20], 이들 모두 클래스별 샘플, 그래디언트, 또는 신뢰할 만한 통계량에 대한 지속적인 접근을 필요로 한다. 대표적으로 CCVR은 업로드된 클래스 통계량에서 샘플링한 가상 특징으로 헤드를 재학습한다 [11]. 프로토타입 기반 분류기는 각 클래스를 특징 공간에서 직접 표현한다는 점에서 특히 매력적이며 [21]–[25], iCaRL은 softmax 헤드를 최근접 프로토타입 판독으로 대체한 대표적 사례이다 [22]. 그러나 차량이 이탈하여 적은 개수로 프로토타입을 구성하면, 실제(live) 표현으로부터 점차 멀어지게 된다.

% 3. 프로토타입 표류 보정
% 이러한 표류를 추적하기 위해 많은 연구가 수행되어 왔다. 일부는 저장된 exemplar로 프로토타입을 재계산하거나 [22] 백본을 동결하여 [24], [25] 표류를 우회하지만, 각각 저장 부담과 가소성 손실을 수반한다. 보상 기반 기법 중 SDC는 가용 표본에서 측정한 임베딩 변위의 가중 평균을 프로토타입에 전이하고 [28], [29], LDC는 구 특징 공간에서 신 특징 공간으로의 사상을 학습해 적용한다 [13], [30], [31]. 그러나 이러한 접근들은 모두 갱신 전후의 두 백본을 통과한 표본 수준(sample-level)의 특징에 의존하며, 이는 원시 데이터가 차량을 떠나지 않는 FL의 프라이버시 제약과 충돌한다 [1], [6], [33]–[35].

\section{Preliminary Study}
\label{sec:prelim}

\begin{figure}[t]
  \centering
  \includegraphics[width=\columnwidth]{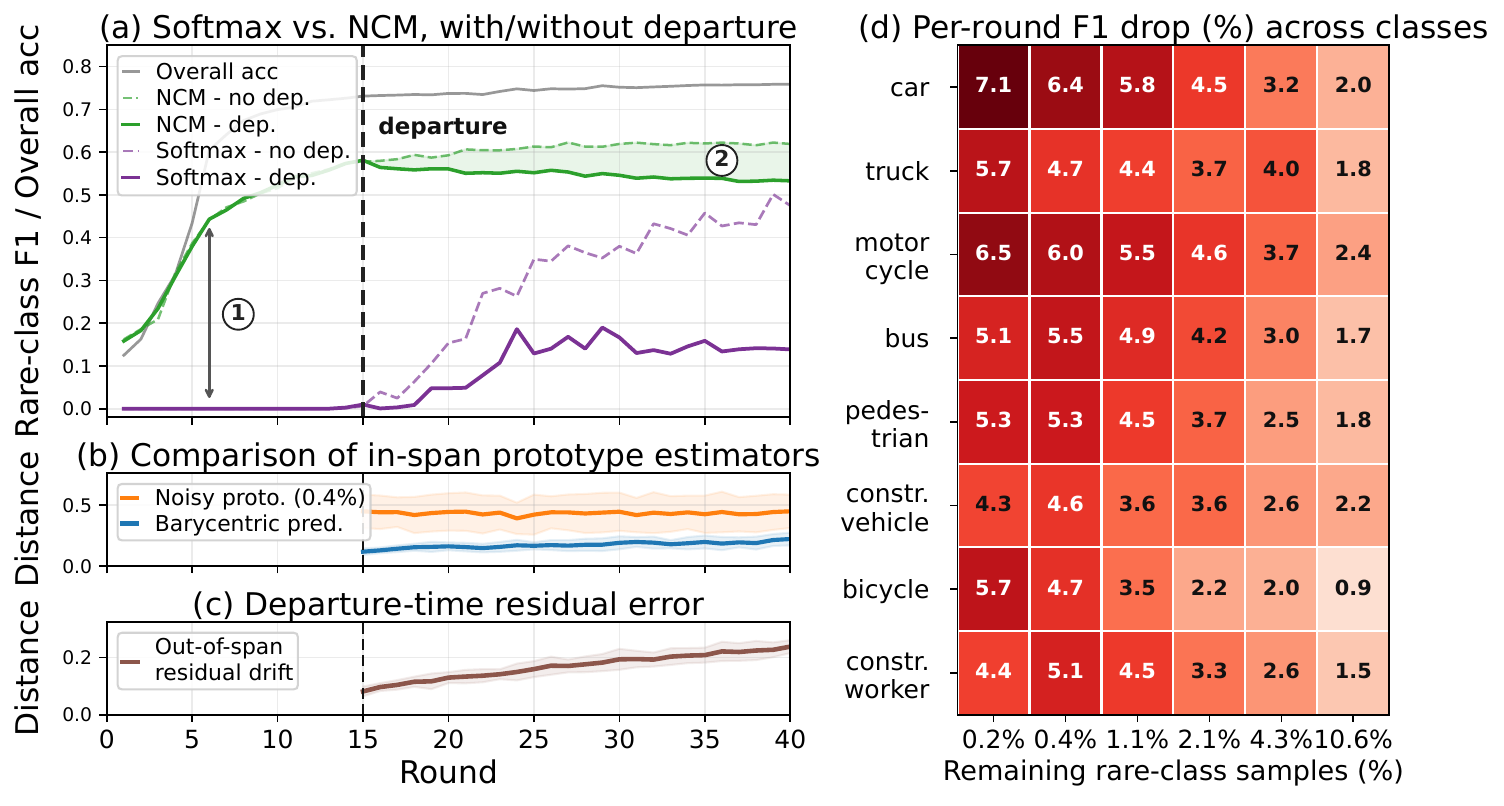}
  \vspace{-3ex}
  \caption{Rare-class performance and prototype drift under vehicle departure. Best viewed in color.} 
  \vspace{-3ex}
  \label{fig:diagnosis}
\end{figure}

We first empirically analyze post-departure rare-class prototype drift and motivate \system{}'s design. As mentioned, cross-vehicle FL requires rapid convergence and continued rare-class recognition. We therefore examine why the prototype of the rare-class becomes misaligned, how class-level statistics can track its in-span and out-of-span drift, and whether the resulting performance loss persists across classes and remaining-sample fractions. In Figure~\ref{fig:diagnosis} we use the nuImages dataset with a three-layer CNN backbone. Figures~\ref{fig:diagnosis}~(a)-(c) use the ``truck'' class as the rare class, assuming that the departing vehicle operates mainly near logistics hubs and holds a majority of the truck samples, with 0.4\% of rare-class samples remaining after its departure at FL round~15. Figure~\ref{fig:diagnosis}~(d) varies both the rare class and the remaining sample fraction.

Figure~\ref{fig:diagnosis}~(a) examines the early-round benefit of prototype-based inference and the effect of vehicle departure. Plots report rare-class F1 and overall accuracy while comparing softmax classification with a prototype-based Nearest Class Mean (NCM) inference approach by replacing the classifier head with nearest-prototype classification. Among the plots with no departure events (dotted lines), NCM reaches high rare-class F1 earlier than softmax (\circled{1}). After departure, however, F1 falls below the non-departure curve (green lines, \circled{2}) and the gap persists. Continued training on the remaining classes causes representation drift, while the few rare-class samples provide only a noisy estimate of its current prototype. As a result, NCM can no longer represent the rare class reliably in such cases, even while overall accuracy continues to rise as performance on the remaining classes improves. Note that this discrepancy suggests that the rare-class F1 score, rather than overall accuracy, serves as a better magnifier of the aforementioned problem. % These results suggest that 
Overall, prototype-based inference improves early-round rare-class performance. Nevertheless, continued FL training after departure requires the rare-class prototype to track representation drift.

To determine how rare-class prototype drift can be tracked under FL privacy constraints, we use the class prototypes reported by remaining clients, which are aggregate statistics rather than raw data or per-sample features. While remaining-class prototypes cannot cover the entire high-dimensional feature space, they naturally define the set of directions that can be represented using the classes still observed after departure. We refer to this set as their \textit{span} and track the rare-class prototype within it by preserving its position relative to the remaining-class prototypes at departure. A natural way to preserve this position is to express it as a \textit{barycentric relationship}, which assigns a weight to each remaining-class prototype according to its contribution to the rare-class location. Applying these weights to the updated remaining-class prototypes then provides the corresponding rare-class position as the feature space updates. 

Figure~\ref{fig:diagnosis}~(b) evaluates this estimate using Euclidean distance to the current oracle in-span component, where a smaller value indicates a more accurate estimate. Making direct estimations from the few remaining rare-class samples yields a large and persistent error, whereas barycentric estimations maintains a smaller and more stable distance to the oracle. This shows that the relative configuration of the remaining classes provides a reliable signal for tracking drift shared across classes. Tracking the in-span component alone, however, cannot ensure the rare-class prototype accuracy.
Figure~\ref{fig:diagnosis}~(c) examines the remaining out-of-span component (i.e., residual) by measuring the distance between the oracle prototype and its in-span component.
The initial residual at departure provides a close estimate, but the mismatch grows steadily as the drift accumulates. Thus, prototype reconstruction must also update the out-of-span residual in addition to accurately track the overall drift.

Finally, Figure~\ref{fig:diagnosis}~(d) examines how broadly the observed post-departure problem extends across classes and remaining-sample conditions. We treat each nuImages class as a rare class (in order) and vary the fraction of the rare class samples remaining after departure. Each value reports the average per-round percentage decrease in rare-class F1 relative to the corresponding non-departure curve over the 25 post-departure rounds marked by \circled{2} in Figure~\ref{fig:diagnosis}~(a). Larger values indicate greater degradation, and classes are ordered in increasing loss order. Loss persists across all classes, increasing as fewer rare-class samples remain, and persists even when samples are relatively available. This suggests that the issue is not confined to a specific class scenario, but arises generally whenever a departing vehicle holds a substantial share of a class.

Together, these results establish the design requirements of \system{}. NCM first provides useful rare-class performance before departure. Afterward, since the few remaining rare-class samples cannot reliably track the changing prototype, \system{} must use updated remaining-class prototypes to estimate the drift shared across classes. Given that their span leaves an out-of-span residual, this too should be updated as representation drift accumulates and calibrated according to the reliability of available rare-class statistics. These requirements motivate the barycentric reconstruction, covariance-based residual prediction, and reliability-aware weighting we discuss in Section~\ref{sec:scheme}.

\section{\system{} Design}
\label{sec:scheme}

\begin{comment}
    
\begin{figure*}[t]
  \centering
  \includegraphics[width=0.6\linewidth]{figures/scheme_overview_v4.pdf}
  \caption{\system{} operations in vehicle departure scenario with barycentric prototype reconstruction and per round drift tracking.}
  \label{fig:overview}
  \vspace{-2ex}
\end{figure*}
\end{comment}

\begin{figure}[t]
  \centering
  \includegraphics[width=\linewidth]{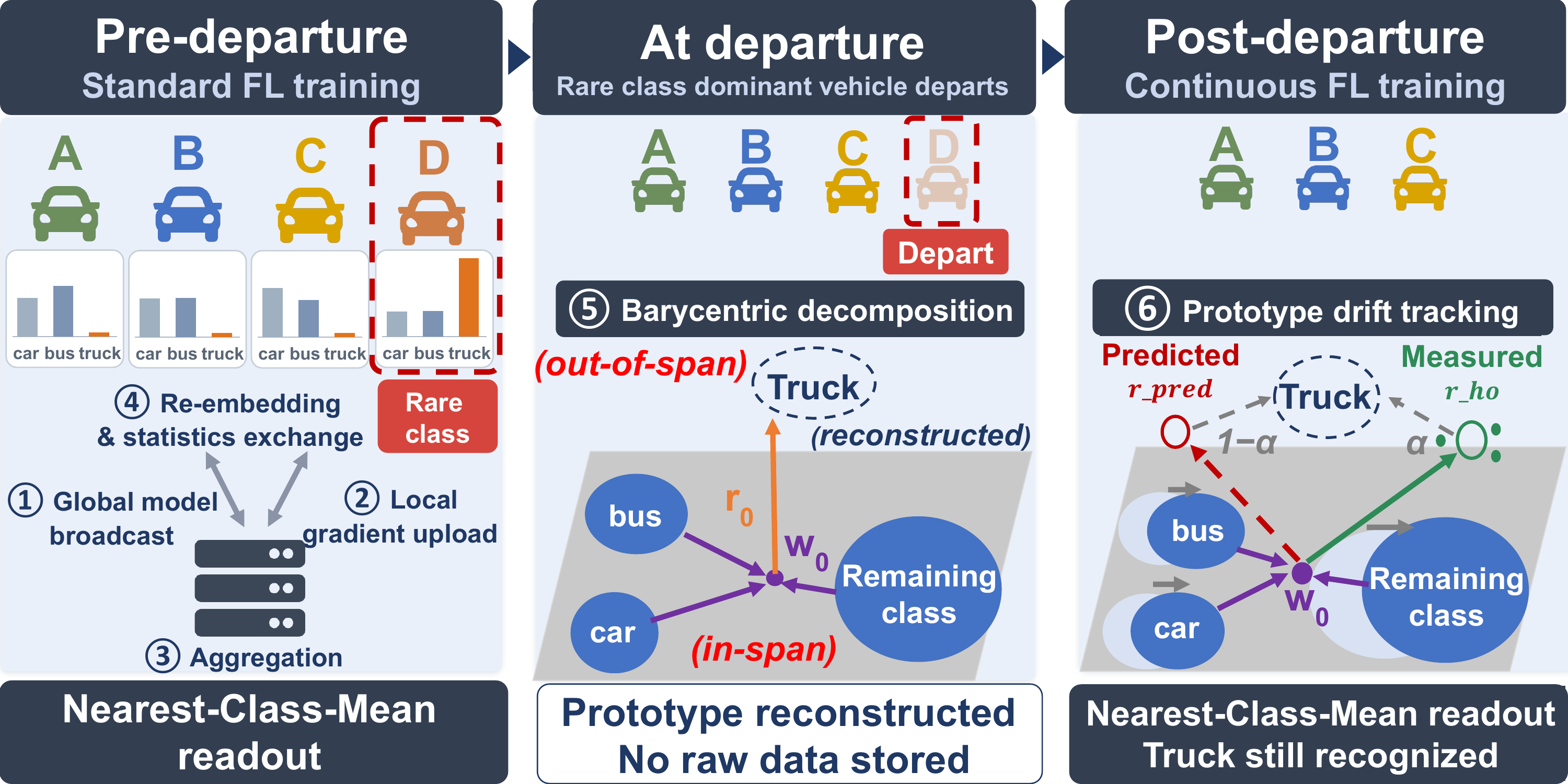}
  \caption{\system{} in vehicle departure scenario with barycentric prototype reconstruction and per-round drift tracking.}
  \label{fig:overview}
  \vspace{-3ex}
\end{figure}

%In this section, we present the problem formulation, core idea, and overall design of \system{}.
Once a vehicle leaves the FL network and only a small number of rare-class samples remain in the federation, the reliable tracking of this rare-class prototype becomes challenging since subsequent model updates can cause representation drift. Prototypes of classes that still remain in the federation can track the drift shared across classes, but cannot capture the changing out-of-span residual. Based on this observation, we formulate this class-dominant vehicle departure scenario to a problem of accurately estimating the rare-class prototype.

% \subsection{Problem Formulation and Core Idea}

% We denote the feature of an input $x$, extracted from the penultimate layer immediately preceding the classifier, by $h_{\theta}(x)\in\mathbb{R}^{d}$, and refer to the mean feature vector of class $c$, denoted by $\bm{\mu}_{c}\in\mathbb{R}^{d}$, as its \textit{prototype}. When a vehicle holding most samples of a rare class departs, the rare-class prototype $\bm{\mu}_{rare}$ can no longer be reliably estimated from the remaining data. Meanwhile, the feature extractor $h_{\theta}(\cdot)$ continues to evolve through the FL rounds on the remaining clients (i.e., vehicles), causing the previously observed prototype to become misaligned with the current feature space. We therefore formulate departure as the problem of estimating the current rare-class prototype under representation drift.

% \system{} addresses this problem by expressing the rare-class prototype as a weighted combination of the pre-departure remaining-class prototypes. Since remaining classes continue to be observed throughout training, their updated prototypes provide reference points for reconstructing the rare-class prototype in the evolving feature space. Any rare-class samples remaining on active clients are used only to calibrate the reconstructed prototype, rather than to update the model. This design relies only on class-level statistics and does not expose raw samples or per-sample features.

\subsection{Problem Formulation}
\label{sec:core-idea}
We denote the feature of an input $x$, extracted from the penultimate layer immediately preceding the classifier, by $h_{\theta}(x)\in\mathbb{R}^{d}$. The prototype of class $c$, $\bm{\mu}_{c}\in\mathbb{R}^{d}$, is the mean feature vector of its samples. \system{} expresses the rare-class prototype $\bm{\mu}_{\mathrm{c}}$ as an \textit{affine combination} of remaining-class prototypes and a residual. After departure, the in-span component is updated using statistics of prototypes from remaining classes, while the out-of-span residual is predicted using covariance changes any available rare-class statistics.

\subsection{Overall Workflow}
\label{sec:overall}
Figure~\ref{fig:overview} illustrates the overall workflow of \system{}. Prior to vehicle departure, \system{} runs standard FL while maintaining globally aligned class-level statistics. At FL round $r$, \circled{1} the server distributes the global model $\theta_r$ to active clients. \circled{2} Each client trains this model using local data and uploads model updates together with local sample counts. \circled{3} The server aggregates updates from active clients to obtain the next global model $\theta_{r+1}$, and \circled{4} $\theta_{r+1}$ is redistributed from the server to active clients for \textit{re-embedding}. With the shared model, clients recompute class-wise feature means, second moments, and sample counts, and returns aggregated statistics to the server. These statistics are aggregated to maintain globally aligned prototypes and shared covariance statistics across all classes. Next, an NCM readout replaces the head (as discussed in $\S$\ref{sec:prelim}) and this procedure is repeated until departure.

As long as the vehicle holding the majority of rare-class samples resides, the rare-class prototype is updated directly from reported class statistics. Immediately before vehicle departure, \circled{5} the server computes and stores a compact relationship weight ($\S$\ref{sec:prototype-estimation}), specifically, the barycentric relationship between the rare-class prototype and the remaining-class prototypes, together with the out-of-span residual. Upon departure, \circled{6} the vehicle no longer contributes to model updates nor provides statistics, but remaining clients continue FL training and report updated class-level statistics. The server reconstructs the current rare-class prototype by applying the stored barycentric relationship to the updated remaining-class prototypes and by predicting residual changes via covariance differences. If a small amount of rare-class samples are available on remaining clients, their aggregate statistics under the current global model can further calibrate prototype reconstruction, which is then inserted into the current prototype set and used by the Mahalanobis NCM classifier.

% \todo{@Hanju and Gyeongmin: I've wrote the overall workflow based on the repository. Please check whether it is technically correct. -`}

\subsection{Rare-Class-Prototype Estimation}
\label{sec:prototype-estimation}
Upon vehicle departure, subsequent FL updates alter the feature space, causing the last observed rare-class prototype to become misaligned with the current representation. To track this drift, \system{} represents the rare-class prototype relative to the continuously observed remaining-class prototypes. At departure, \system{} decomposes the rare-class prototype into an affine combination of the remaining-class prototypes and a residual. In future FL rounds, \system{} applies the fixed affine weights to the updated prototypes and adapts a measured and predicted residual to the current feature-space geometry.

\subsubsection{Departure-time prototype decomposition}

Immediately before the rare-class-dominant vehicle leaves, the server observes prototype $\bm{\mu}_c\in\mathbb{R}^{d}$ for rare class $c$ at departure round $t_{\mathrm{d}}$. Let $\mathbf{M}_{\mathrm{d}}\in\mathbb{R}^{K\times d}$ denote the matrix whose rows are prototypes of the $K$ remaining classes. \system{} identifies their affine combination that best approximates $\bm{\mu}_c$:
\begin{equation}
\mathbf{w}_0
=
\operatorname*{arg\,min}_{\substack{
    \mathbf{w}\in\mathbb{R}^{K} \\
    \mathbf{1}^{\top}\mathbf{w}=1
}}
\left\{
    \left\|
        \bm{\mu}_c-\mathbf{M}_{\mathrm{d}}^{\top}\mathbf{w}
    \right\|_2^2
    +
    \lambda\left\|\mathbf{w}\right\|_2^2
\right\}.
\label{eq:deposit_w}
\end{equation}

Here, $\mathbf{M}_{\mathrm{d}}^{\top}\mathbf{w}\in\mathbb{R}^{d}$ is the affine combination specified by $\mathbf{w}$. Equation~\ref{eq:deposit_w} minimizes its discrepancy from the rare-class prototype. Each element of $\mathbf{w}_0$ specifies the coefficient assigned to corresponding remaining-class prototypes. Constraint $\mathbf{1}^{\top}\mathbf{w}=1$ makes the relationship invariant to a common translation of prototypes, while the regularization term $\lambda\left\|\mathbf{w}\right\|_2^2$ discourages excessively large coefficients and stabilizes the solution. Furthermore, since the affine combination may not fully recover $\bm{\mu}_c$, \system{} stores the remaining residual as
\begin{equation}
\mathbf{r}_0
=
\bm{\mu}_c
-
\mathbf{M}_{\mathrm{d}}^{\top}\mathbf{w}_0.
\label{eq:deposit_r}
\end{equation}

The departure-time rare-class prototype can thus be written as $\bm{\mu}_c=\mathbf{M}_{\mathrm{d}}^{\top}\mathbf{w}_0+\mathbf{r}_0$, where $\mathbf{r}_0$ captures the component unexplained by the remaining-class prototypes. However, $\mathbf{r}_0$ is defined in departure-time feature space, whose directional scales and correlations may change as FL continues. Therefore, directly reusing $\mathbf{r}_0$ would ignore the geometric effects of representation drift.
To characterize this geometry, \system{} stores the pooled within-class covariance of remaining-class features. Since these classes remain observable, their covariance can be used to track changes in the feature space. To avoid numerical instability caused by a singular or poorly conditioned empirical covariance, \system{} applies trace-scaled ridge regularization:
\begin{equation}
\bar{\sigma}_{\mathrm{d}}
=
\frac{1}{d}
\operatorname{tr}\left(\bm{\Sigma}_{\mathrm{d}}\right), \;\;
\tilde{\bm{\Sigma}}_{\mathrm{d}}
=
\bm{\Sigma}_{\mathrm{d}}
+
\lambda_{\Sigma}
\bar{\sigma}_{\mathrm{d}}
\mathbf{I}_{d}.
\label{eq:deposit_cov}
\end{equation}

Here, $\bm{\Sigma}_{\mathrm{d}}\in\mathbb{R}^{d\times d}$ is the pooled within-class covariance, $\bar{\sigma}_{\mathrm{d}}$ is the average marginal variance, and $\lambda_{\Sigma}>0$ controls the regularization strength. The added isotropic term increases all eigenvalues by $\lambda_{\Sigma}\bar{\sigma}_{\mathrm{d}}$, reducing sensitivity to poorly estimated low-variance directions. Scaling it by the average variance makes the regularization adaptive to the overall feature scale. Upon vehicle departure, \system{} computes the corresponding regularized covariance from the remaining classes. The change from $\tilde{\bm{\Sigma}}_{\mathrm{d}}$ captures the feature-space geometry evolution and is used to transform $\mathbf{r}_0$ before combining it with the updated in-span component. Thus, the in-span component follows the movement of remaining-class prototypes, while covariance-based prediction adapts the unexplained residual.

\subsubsection{Post-departure calibration}
\label{subsec:postdep-predcal}
%As the class-dominant vehicle departs, remaining clients continue training using their local data, causing the feature space to evolve across subsequent FL rounds. However, once departed, the oracle rare-class prototype that would be obtained if sufficient class-$c$ data remained available for training is no longer observable. 

Given the class-$c$ prototype drift and continuous changes in feature space with subsequent FL operations after the class-dominant vehicle's departure, \system{} makes estimates of class-$c$'s prototype at each round $t>t_{\mathrm{d}}$. We model this reconstruction as the sum of a shared movement tracked by the remaining classes and a residual that is either directly measured or predicted:
\begin{equation}
\hat{\bm{\mu}}_c(t)
=
\underbrace{
\mathbf{M}_t^{\top}\mathbf{w}_0
}_{\text{shared drift}}
+
\underbrace{
\alpha_t\hat{\mathbf{r}}_{\mathrm{ho}}(t)
}_{\text{measured residual}}
+
\underbrace{
\left(1-\alpha_t\right)
\hat{\mathbf{r}}_{\mathrm{pred}}(t)
}_{\text{predicted residual}},
\label{eq:assembly}
\end{equation}
The first term applies the departure-time affine weights to the current remaining-class prototypes, allowing the reconstructed prototype to follow feature-space changes shared across classes. The second term directly measures the residual from any class-$c$ samples still available on remaining vehicles. When these samples are insufficient, the third term predicts the residual by transporting the stored departure-time residual according to the observed covariance change of remaining-class features. Coefficient $\alpha_t$ balances the measured and predicted residuals based on the reliability of available class-$c$ samples.

\paragraph{Shared drift from the remaining classes}
The first term in Equation~\ref{eq:assembly} applies the departure-time weights $\mathbf{w}_0$ to remaining-class prototypes $\mathbf{M}_t$. As these prototypes are updated, $\mathbf{M}_t^{\top}\mathbf{w}_0$ follows representation changes (i.e., drift) shared across classes. \system{} keeps $\mathbf{w}_0$ fixed since re-estimates require the unavailable oracle rare-class prototype. Thus, the relative relationship at $t_d$ is preserved, while the prototype to which it is applied continues to evolve.

\paragraph{Sample-based measured residual}
Since the affine component may not reflect rare-class specific drift, \system{} uses remaining rare-class samples to calibrate the reconstructed prototype. Let $\mathcal{D}_{c}^{\mathrm{ho}}$ denote remaining rare-class samples, which are held out and not used for backbone training, and $\mathbf{P}_t$ be the orthogonal projector onto the subspace spanned by the current remaining-class prototypes in $\mathbf{M}_t$. Its complementary projector, $\mathbf{P}_t^{\perp}=\mathbf{I}_d-\mathbf{P}_t$, isolates feature components that cannot be represented by remaining prototypes. \system{} measures the residual by projecting the mean feature of the remaining class-$c$ samples onto this complementary subspace:
\begin{equation}
\hat{\mathbf{r}}_{\mathrm{ho}}(t)
=
\mathbf{P}_t^{\perp}
\mathbb{E}_{x\in\mathcal{D}_{c}^{\mathrm{ho}}}
\left[
h_{\theta_t}(x)
\right],
\label{eq:measured_residual}
\end{equation}
This measured residual directly captures the rare-class specific representation drift that is not explained by using the remaining-class prototypes. However, we note that its reliability can still decrease (or be limited) when only a small number of rare-class samples remain.

\paragraph{Covariance-based predicted residual}
To address the limitation of limited class $c$ samples, \system{} also makes predictions of the residual using the evolving remaining-class feature geometry. Using departure-time covariance $\tilde{\bm{\Sigma}}_{\mathrm{d}}$ and current counterpart $\tilde{\bm{\Sigma}}_t$, \system{} computes the Bures-Wasserstein transport map~\cite{bhatia2019bures} to apply to the stored residual:
\begin{equation}
\begin{aligned}
\mathbf{T}_t
&=
\tilde{\bm{\Sigma}}_{\mathrm{d}}^{-1/2}
\left(
\tilde{\bm{\Sigma}}_{\mathrm{d}}^{1/2}
\tilde{\bm{\Sigma}}_t
\tilde{\bm{\Sigma}}_{\mathrm{d}}^{1/2}
\right)^{1/2}
\tilde{\bm{\Sigma}}_{\mathrm{d}}^{-1/2}, \\
\hat{\mathbf{r}}_{\mathrm{pred}}(t)
&=
\mathbf{P}_t^{\perp}
\mathbf{T}_t
\mathbf{r}_0.
\end{aligned}
\label{eq:residual_transport}
\end{equation}
The transport map $\mathbf{T}_t$ adapts the scale and orientation of $\mathbf{r}_0$ to the current feature geometry. Since it relies only on remaining-class covariances, $\hat{\mathbf{r}}_{\mathrm{pred}}(t)$ can be obtained even without any rare-class samples. We note that this transformation can be replaced with a single scalar by scaling the residual with respect to the out-of-span component's mean magnitude after departure. This change can reduce the communication cost from a $d\times d$ matrix to a single growth factor; making it especially effective for linear backbone feature spaces. We later refer to this configuration as \textit{\system{}-lite} in our evaluations.

\paragraph{Adaptive residual weighting}
\system{} combines the two residuals according to the reliability of remaining rare-class samples. Let $N_t=\left|\mathcal{D}_{c}^{\mathrm{ho}}\right|$ and let $\hat{\sigma}_t^2$ denote the estimated per-sample variance of the measured residual. We compute
\begin{equation}
\begin{aligned}
D_t
&=
\frac{1}{d}
\left\|
\hat{\mathbf{r}}_{\mathrm{ho}}(t)
-
\hat{\mathbf{r}}_{\mathrm{pred}}(t)
\right\|_2^2, \\
\hat{\tau}_t^2
&=
\max\left(
D_t-\frac{\hat{\sigma}_t^2}{N_t},
\varepsilon
\right),
~~\alpha_t
=
\frac{\hat{\tau}_t^2}
{\hat{\tau}_t^2+\hat{\sigma}_t^2/N_t},
\end{aligned}
\label{eq:alpha}
\end{equation}
where $\varepsilon>0$ ensures numerical stability. Larger and reliable rare-class samples increase $\alpha_t$, assigning greater weight to measurements. Otherwise, the reconstruction relies more on the covariance-based prediction. When $N_t=0$, we set $\alpha_t=0$.

\subsection{Mahalanobis Nearest-Class-Mean Readout}
\label{sec:mahalanobis}
At round $t$, \system{} forms a prototype set consisting of the reconstructed rare-class prototype and the current prototypes of the remaining classes, and predicts the class with prototype closest to the input under a regularized Mahalanobis distance~\cite{mclachlan1999mahalanobis}, with the covariance estimated from the remaining-class features as in Equation~\ref{eq:deposit_cov}. This nearest-class-mean (NCM) readout~\cite{mensink2013distance} incorporates the estimated rare-class representation without retraining the model, and avoids the original classifier which becomes increasingly biased after departure. The Mahalanobis metric accounts for direction-dependent feature variation, which helps distinguish the rare class from nearby confusable classes while reusing the covariance statistics already maintained by \system{}.

\section{Evaluation}\label{sec:eval}
We now evaluate \system{} using extensive experiments with three datasets and various backbones in a vehicular FL setting.
 
\subsection{Evaluation Setup}
\noindent\textbf{Datasets.}
We evaluate \system{} on two road-object classification datasets, nuImages~\cite{caesar2020nuscenes} and Car-1000~\cite{hu2025car}. We use nuImages as the primary dataset, comprising 9.3K road-scene images annotated with eight object classes, including cars, pedestrians, and trucks. Using the provided bounding-box annotations, we crop each object from the original scene and treat the resulting crop as an individual sample for classification. Car-1000 contains 140K images organized into seven coarse-grained vehicle categories and 1,000 fine-grained vehicle models. The coarse-grained categories include SUVs, sedans, vans, and buses. To construct a controlled FL setting with balanced global class frequencies, we sample up to 1,000 images from each class. 

\vspace{0.5ex}

\noindent\textbf{Classifier Backbones.}
To evaluate the robustness of \system{} across different backbone architectures, we consider three CNN architectures and one vision transformer. CNNs remain widely used for vision-based tasks, while the transformer model evaluates whether \system{} generalizes beyond architectures. Our default backbone is a lightweight three-layer CNN composed of cascaded convolution and batch-normalization layers, similar to VGG~\cite{simonyan2014very}, with a feature dimension of $d=128$. We additionally evaluate ResNet-18 ($d=512$)~\cite{he2016deep}, MobileNetV3 ($d=1280$)~\cite{howard2019searching}, and ViT-Tiny ($d=192$)~\cite{dosovitskiy2020image}.
\begin{table}[t]
\centering
\scriptsize
\setlength{\tabcolsep}{3.0pt}
\renewcommand{\arraystretch}{1.06}
\begin{adjustbox}{max width=\columnwidth}
\begin{tabular}{@{}lllcccc@{}}
\toprule
Category & Mechanism & Method & PD & HFL & PF & TF \\
\midrule
Standard FL
& Aggregation
& FedAvg~\cite{mcmahan2017communication}
& \textcolor{red}{\xmark}
& \textcolor{blue}{\cmark}
& \textcolor{blue}{\cmark}
& \textcolor{blue}{\cmark} \\
\addlinespace[1.5pt]
FL calibration
& Virtual features
& CCVR~\cite{luo2021no}
& \textcolor{red}{\xmark}
& \textcolor{blue}{\cmark}
& \textcolor{blue}{\cmark}
& \textcolor{red}{\xmark} \\
\addlinespace[1.5pt]
Prototype re-est.
& Sample mean
& iCaRL-NME~\cite{rebuffi2017icarl}
& \textcolor{red}{\xmark}
& \textcolor{red}{\xmark}
& \textcolor{red}{\xmark}
& \textcolor{blue}{\cmark} \\
\addlinespace[1.5pt]
\multirow{2}{*}{Drift compensation}
& Additive drift
& SDC~\cite{yu2020semantic}
& \textcolor{blue}{\cmark}
& \textcolor{red}{\xmark}
& \textcolor{red}{\xmark}
& \textcolor{blue}{\cmark} \\
& Learned mapping
& LDC~\cite{gomez2024exemplar}
& \textcolor{blue}{\cmark}
& \textcolor{red}{\xmark}
& \textcolor{red}{\xmark}
& \textcolor{red}{\xmark} \\
\addlinespace[1.5pt]
\multirow{2}{*}{Reference bounds}
& Frozen prototype
& Frozen
& \textcolor{red}{\xmark}
& \textcolor{blue}{\cmark}
& \textcolor{blue}{\cmark}
& \textcolor{blue}{\cmark} \\
& Test dataset
& Oracle
& \textcolor{blue}{\cmark}
& \textcolor{red}{\xmark}
& \textcolor{red}{\xmark}
& \textcolor{blue}{\cmark} \\
\addlinespace[1.5pt]
Proposed
& Reconstruction
& \textbf{\system{}}
& \textcolor{blue}{\cmark}
& \textcolor{blue}{\cmark}
& \textcolor{blue}{\cmark}
& \textcolor{blue}{\cmark} \\
\bottomrule
\end{tabular}
\end{adjustbox}
\caption{Capability comparison under rare-class vehicle departure. PD denotes post-departure drift compensation; HFL, heterogeneous FL support; PF, no per-sample feature storage; and TF, no additional training.}
\label{tab:baseline-capabilities}
\vspace{-3ex}
\end{table}

\vspace{0.5ex}
\noindent\textbf{Scenario.}
We evaluate \system{} over 100 FL rounds with 5 local epochs in a cross-vehicle departure scenario. Unless otherwise stated, the rare-class-dominant vehicle departs at round 15 and initially holds 98\% of the rare-class samples, creating a highly non-IID class distribution. The remaining 2\% of the rare-class samples are distributed across the active vehicles, while samples from all other classes are distributed uniformly.

For nuImages, we designate \emph{truck} as the rare class, reflecting a scenario in which a vehicle operating near logistics hubs collects most of the truck samples, as illustrated in Figure~\ref{fig:scope}. For Car-1000, we use its seven coarse-grained vehicle categories and designate \emph{van} as the rare class. We also construct a more challenging fine-grained setting using sedan categories grouped by vehicle size, with \emph{micro sedan} designated as the rare class. This setting evaluates \system{} when the rare and remaining classes exhibit highly similar visual characteristics.

% \noindent\textbf{Scenario.}
% We evaluate \system{} for 100 FL rounds, with the rare class dominant vehicle departing at round $15$. We note that changing the departure round has no effect on the results. The federation has as many vehicles as class numbers, and every class contributes 1,000 samples in total. All classes except the rare class are split evenly among vehicles, while the rare class dominant vehicle holds $n \in \{90\%, 95\%, 97.5\%, 99\%\}$ of the rare class; leaving only $\{100, 50, 25, 10\}$ rare class samples (among 1,000) to remain across remaining clients after departure.

% For nuImages, we designate \emph{truck} as the rare class, reflecting a scenario in which the departing vehicle operates near logistics hubs and collects most truck samples, as illustrated in Fig.~\ref{fig:scope}. For Car-1000, we use seven coarse-grained vehicle categories and designate \emph{sports car} as the rare class. We further construct a more challenging fine-grained setting by classifying sedan subcategories according to vehicle size, with \emph{micro sedan} as the rare class. This setting evaluates whether \system{} can preserve rare-class recognition when the rare and remaining classes exhibit highly similar visual features.

\vspace{0.5ex}
\noindent\textbf{Baselines.}
Table~\ref{tab:baseline-capabilities} compares \system{} (and \system{}-lite for some experiments) with five baseline categories that represent standard FL, classifier calibration, direct prototype re-estimation, explicit drift compensation, and reference performance bounds.

\begin{itemize}[leftmargin=*]
    \item \textbf{Standard FL.}
    FedAvg~\cite{mcmahan2017communication} continues training the shared backbone and softmax classifier after departure without correcting rare-class representation. This represents a standard FL reference for quantifying departure-induced degradation.

    \item \textbf{FL calibration.}
    CCVR~\cite{luo2021no} generates virtual features from aggregated class-level statistics and retrains the classifier to reduce classifier bias under heterogeneous FL. This is a relevant comparison since it uses privacy-preserving class statistics similar to \system{}, but does not compensate for rare-class prototype drift as the backbone evolves.

    \item \textbf{Prototype re-estimation.}
    iCaRL-NME~\cite{rebuffi2017icarl} directly recomputes the rare-class prototype from the limited samples remaining after departure and performs nearest-mean classification. This baseline isolates whether direct prototype estimation from scarce rare-class data is sufficient, without explicitly modeling representation drift.

    \item \textbf{Drift compensation.}
    SDC~\cite{yu2020semantic} estimates an additive prototype drift from paired feature drift, whereas LDC~\cite{gomez2024exemplar} learns a mapping between successive features. These methods represent explicit prototype-drift compensation, but require sample-level feature pairs, and LDC additionally requires training a mapping model. We use both methods using the same post-departure rare-class sample budget as \system{}.

    \item \textbf{Reference bounds.}
    \emph{Frozen Prototype} reuses the departure-time rare-class prototype without drift correction, showing the effect of prototype staleness. \emph{Oracle} recomputes the prototype using test rare-class data embedded by the current backbone, providing an upper reference that is unavailable in the actual departure scenario.

\end{itemize}

\vspace{0.5ex}
\noindent\textbf{Evaluation Metric and Environment.}
As shown in Section~\ref{sec:prelim} (c.f., Fig.~\ref{fig:diagnosis} (a)), overall accuracy can remain high even when rare-class performance drops. We therefore use rare-class F1 score as our primary metric, given that it considers both recall and precision. We conduct all experiments on a server equipped with four NVIDIA GeForce RTX 2080 Ti GPUs with 11~GB of memory each and an Intel Xeon Silver 4210 CPU with 10 cores and 20 threads running at 2.20~GHz.

\subsection{Overall Performance}
\label{sec:eval-overall}

\begin{figure}[t]
    \centering
    \includegraphics[width=\columnwidth]{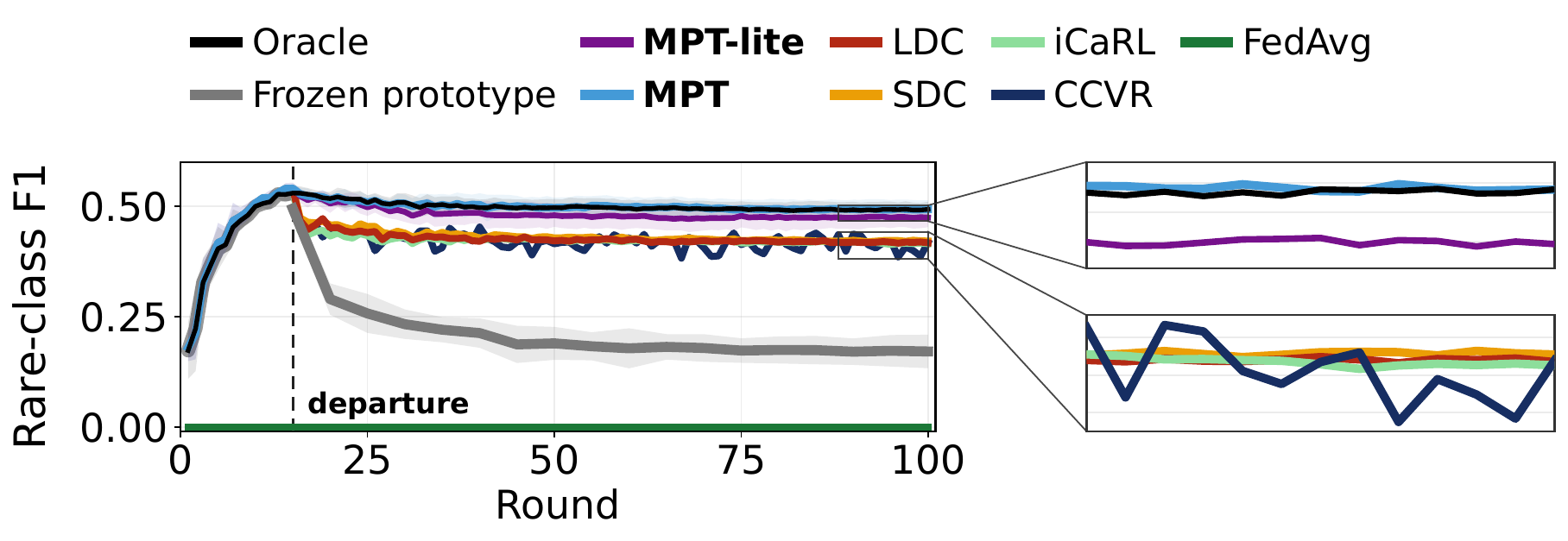}
    \caption{Per-round rare-class F1 before and after vehicle departure on nuImages using the three-layer CNN backbone, with 2\% of the rare-class samples remaining. Best viewed in color.}
    \label{fig:eval_per_round}
    \vspace{-2ex}
\end{figure}

\begin{table*}[t]
\centering
\small
\setlength{\tabcolsep}{3pt}
\resizebox{\textwidth}{!}{%
\begin{tabular}{l cccc cccc cccc}
\toprule
 & \multicolumn{4}{c}{nuImages (Truck)} & \multicolumn{4}{c}{Coarse-grained Car-1000 (Van)} & \multicolumn{4}{c}{Fine-grained Car-1000 (Micro Sedan)} \\
\cmidrule(lr){2-5}\cmidrule(lr){6-9}\cmidrule(lr){10-13}
Method & $1\%$ & $2\%$ & $5\%$ & $10\%$ & $1\%$ & $2\%$ & $5\%$ & $10\%$ & $1\%$ & $2\%$ & $5\%$ & $10\%$ \\
\midrule
CCVR & .384{\scriptsize$\pm$.031} & .433{\scriptsize$\pm$.018} & .478{\scriptsize$\pm$.017} & .479{\scriptsize$\pm$.024} & .360{\scriptsize$\pm$.033} & .382{\scriptsize$\pm$.020} & .413{\scriptsize$\pm$.006} & .426{\scriptsize$\pm$.022} & .213{\scriptsize$\pm$.035} & .234{\scriptsize$\pm$.011} & .254{\scriptsize$\pm$.027} & .268{\scriptsize$\pm$.012} \\
iCaRL & .393{\scriptsize$\pm$.014} & .419{\scriptsize$\pm$.004} & .459{\scriptsize$\pm$.019} & .434{\scriptsize$\pm$.022} & .357{\scriptsize$\pm$.034} & .364{\scriptsize$\pm$.016} & .392{\scriptsize$\pm$.011} & .403{\scriptsize$\pm$.016} & .209{\scriptsize$\pm$.029} & .224{\scriptsize$\pm$.012} & .251{\scriptsize$\pm$.035} & .258{\scriptsize$\pm$.013} \\
SDC & .435{\scriptsize$\pm$.023} & .436{\scriptsize$\pm$.010} & .456{\scriptsize$\pm$.008} & .451{\scriptsize$\pm$.018} & .417{\scriptsize$\pm$.019} & .422{\scriptsize$\pm$.018} & .427{\scriptsize$\pm$.002} & .426{\scriptsize$\pm$.018} & .213{\scriptsize$\pm$.038} & .232{\scriptsize$\pm$.019} & .249{\scriptsize$\pm$.023} & .253{\scriptsize$\pm$.026} \\
LDC & .421{\scriptsize$\pm$.012} & .425{\scriptsize$\pm$.017} & .457{\scriptsize$\pm$.007} & .451{\scriptsize$\pm$.017} & .388{\scriptsize$\pm$.015} & .412{\scriptsize$\pm$.028} & .425{\scriptsize$\pm$.007} & .425{\scriptsize$\pm$.017} & .246{\scriptsize$\pm$.024} & .283{\scriptsize$\pm$.008} & .289{\scriptsize$\pm$.022} & .283{\scriptsize$\pm$.021} \\
Frozen prototype & .186{\scriptsize$\pm$.042} & .210{\scriptsize$\pm$.045} & .199{\scriptsize$\pm$.043} & .202{\scriptsize$\pm$.046} & .155{\scriptsize$\pm$.030} & .145{\scriptsize$\pm$.026} & .135{\scriptsize$\pm$.009} & .184{\scriptsize$\pm$.029} & .088{\scriptsize$\pm$.016} & .061{\scriptsize$\pm$.001} & .105{\scriptsize$\pm$.006} & .058{\scriptsize$\pm$.005} \\
\system{}-lite & .507{\scriptsize$\pm$.012} & .501{\scriptsize$\pm$.010} & .510{\scriptsize$\pm$.001} & \textbf{.512{\scriptsize$\pm$.020}} & .429{\scriptsize$\pm$.005} & .429{\scriptsize$\pm$.005} & .433{\scriptsize$\pm$.014} & .437{\scriptsize$\pm$.018} & .191{\scriptsize$\pm$.028} & .205{\scriptsize$\pm$.054} & .279{\scriptsize$\pm$.032} & .192{\scriptsize$\pm$.114} \\ 
\textbf{\system{}} & \textbf{.516{\scriptsize$\pm$.009}} & \textbf{.510{\scriptsize$\pm$.018}} & \textbf{.513{\scriptsize$\pm$.005}} & .509{\scriptsize$\pm$.017} & \textbf{.449{\scriptsize$\pm$.017}} & \textbf{.439{\scriptsize$\pm$.017}} & \textbf{.437{\scriptsize$\pm$.009}} & \textbf{.444{\scriptsize$\pm$.022}} & \textbf{.295{\scriptsize$\pm$.032}} & \textbf{.331{\scriptsize$\pm$.007}} & \textbf{.347{\scriptsize$\pm$.030}} & \textbf{.338{\scriptsize$\pm$.008}} \\
\cmidrule(l){1-13}
\textcolor{gray}{Oracle prototype} & \textcolor{gray}{.505{\scriptsize$\pm$.009}} & \textcolor{gray}{.510{\scriptsize$\pm$.012}} & \textcolor{gray}{.509{\scriptsize$\pm$.005}} & \textcolor{gray}{.507{\scriptsize$\pm$.020}} & \textcolor{gray}{.439{\scriptsize$\pm$.013}} & \textcolor{gray}{.441{\scriptsize$\pm$.018}} & \textcolor{gray}{.430{\scriptsize$\pm$.006}} & \textcolor{gray}{.425{\scriptsize$\pm$.023}} & \textcolor{gray}{.315{\scriptsize$\pm$.012}} & \textcolor{gray}{.331{\scriptsize$\pm$.007}} & \textcolor{gray}{.320{\scriptsize$\pm$.013}} & \textcolor{gray}{.322{\scriptsize$\pm$.017}} \\
\bottomrule
\end{tabular}}
\caption{Rare-class F1 for different datasets using CNN backbone, $R=100$.}
\label{tab:dataset_var}
 \vspace{-3ex}
\end{table*}

We first evaluate the overall effectiveness of \system{} by comparing its rare-class F1 against representative baselines and the oracle under a vehicle departure scenario. We use the default setting where all experiments use nuImages with the 3-layer CNN backbone, vehicle departure at round~15, and 2\% of the rare-class samples remaining across the active client pool. We compare \system{} and \system{}-lite to identify when scalar residual scaling is sufficient and when \system{} is needed. %We compare two alternatives of our \system{}, the full and the lite version.

Figure~\ref{fig:eval_per_round} reports per-round rare-class F1 before and after departure. Both \system{} and \system{}-lite remain closest to the oracle, especially \system{}, which nearly overlaps throughout the post-departure rounds. Before departure, FedAvg fails to learn the rare class due to classifier-head bias under the highly skewed distribution, whereas prototype-based methods provide stable rare-class recognition. After departure, Frozen prototype, which retains the departure-time rare-class prototype, progressively degrades as backbone updates gradually misalign with the current feature space. These observations confirm the pre- and post-departure challenges identified in Section~\ref{sec:prelim}. The remaining baselines maintain reasonable performance, including CCVR (heterogeneous FL calibration), iCaRL (direct prototype re-estimation), and SDC and LDC (drift compensation), but all remain below \system{}. These results show that tracking shared drift and updating the out-of-span residual allows \system{} to keep the rare-class prototype closely aligned with the evolving backbone.

%%%%%%%%%%%%%%%%%%%%%%%%%%%%%%%%%%%

\begin{figure}[t]
    \centering
    \includegraphics[width=\columnwidth]{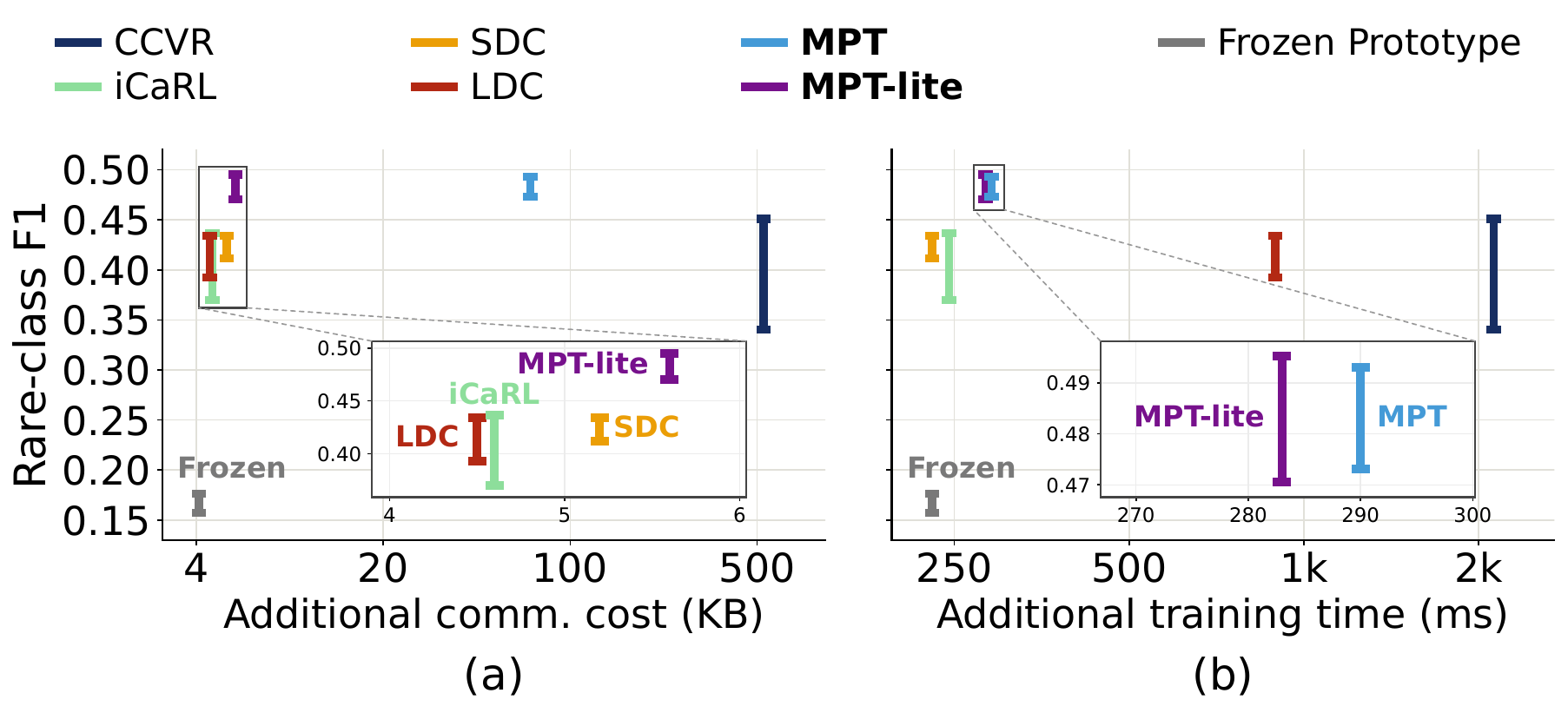}
    \caption{Per-round additional overhead over FedAvg against rare-class F1. (nuImages, CNN, $R=100$). (a) Communication overhead. (b) Wall-clock training overhead. Bars span F1 over the remaining rare-class fraction.}
    \label{fig:eval_comm_tradeoff}
 \vspace{-3ex}
\end{figure}

Next, we evaluate the system costs of each method together with its rare-class F1, focusing on the additional communication and per-round training time. Figure~\ref{fig:eval_comm_tradeoff} compares rare-class F1 at round~100 against the overhead relative to FedAvg, using communication cost in Figure~\ref{fig:eval_comm_tradeoff}~(a) and wall-clock time overhead in Figure~\ref{fig:eval_comm_tradeoff}~(b). Thus, FedAvg shows no overhead and methods positioned upper-left provide a more favorable balance between rare-class recognition and FL system cost.
% The inset enlarges the low-cost cluster.

Figure~\ref{fig:eval_comm_tradeoff}~(a) examines the communication cost of each approach. The increase from FedAvg to the Frozen Prototype case reflects the prototype exchange overhead required by all prototype-based methods, while smaller differences among Frozen Prototype, iCaRL, LDC, and SDC arise from their method-specific requirements. Within this group, \system{}-lite achieves a larger F1 gain with only a small increase in cost. Its additional payload consists of \(K\) barycentric coefficients, where \(K\) is the number of remaining classes, a \(d\)-dimensional stored residual, where \(d\) is the feature dimension, and one scalar growth factor, for \(K+d+1\) scalars in total. This compact information is sufficient to track the shared drift and update the out-of-span residual without transmitting covariance matrices. The (more noticeable) cost increase from \system{}-lite to \system{} comes from transmitting one covariance summary per client for constructing the pooled covariance in Eq.~(\ref{eq:residual_transport}). CCVR incurs a further increase by transmitting a separate covariance for each class at all clients to model class-wise feature distributions, sample ghost features at the server, and retrain the classifier head. In contrast, \system{} uses a single covariance to capture changes in the overall feature space and adapt the out-of-span residual; achieving higher rare-class F1 with substantially less covariance matrices exchange.

We examine the additional wall-clock training time introduced in each FL round using Figure~\ref{fig:eval_comm_tradeoff}~(b). \system{} and \system{}-lite incur nearly identical overhead, showing that covariance-based residual correction increases communication payload while adding negligible computation costs. SDC and iCaRL are slightly faster since they mainly aggregate or update the required statistics without auxiliary model training, but their lower rare-class F1 gives a less favorable time-performance trade-off. LDC and CCVR incur substantially larger overhead due to drift-map training in LDC and ghost-sample generation followed by classifier-head retraining in CCVR. The wider bars of these baselines also show that their performance varies more with the number of remaining rare-class samples. In comparison, \system{} maintains a stable F1 through the adaptive residual weighting in Section~\ref{subsec:postdep-predcal}. Overall, \system{}-lite offers the lower-communication option, while \system{} achieves the strongest rare-class F1 at nearly the same overhead.

\begin{table*}[t]
\centering
\small
\setlength{\tabcolsep}{3pt}
\resizebox{\textwidth}{!}{%
\begin{tabular}{l cccc cccc cccc cccc}
\toprule
 & \multicolumn{4}{c}{Three-layer CNN} & \multicolumn{4}{c}{ResNet-18} & \multicolumn{4}{c}{MobileNetV3} & \multicolumn{4}{c}{ViT-Tiny} \\
\cmidrule(lr){2-5}\cmidrule(lr){6-9}\cmidrule(lr){10-13}\cmidrule(lr){14-17}
Method & $1\%$ & $2\%$ & $5\%$ & $10\%$ & $1\%$ & $2\%$ & $5\%$ & $10\%$ & $1\%$ & $2\%$ & $5\%$ & $10\%$ & $1\%$ & $2\%$ & $5\%$ & $10\%$ \\
\midrule
CCVR & .384{\scriptsize$\pm$.031} & .433{\scriptsize$\pm$.018} & .478{\scriptsize$\pm$.017} & .479{\scriptsize$\pm$.024} & .319{\scriptsize$\pm$.030} & .417{\scriptsize$\pm$.032} & .446{\scriptsize$\pm$.010} & .453{\scriptsize$\pm$.014} & .222{\scriptsize$\pm$.044} & .315{\scriptsize$\pm$.013} & .321{\scriptsize$\pm$.024} & .326{\scriptsize$\pm$.010} & .069{\scriptsize$\pm$.013} & .140{\scriptsize$\pm$.015} & .186{\scriptsize$\pm$.023} & .193{\scriptsize$\pm$.015} \\
iCaRL & .393{\scriptsize$\pm$.014} & .419{\scriptsize$\pm$.004} & .459{\scriptsize$\pm$.019} & .434{\scriptsize$\pm$.022} & .402{\scriptsize$\pm$.007} & .414{\scriptsize$\pm$.022} & .425{\scriptsize$\pm$.001} & .405{\scriptsize$\pm$.013} & .208{\scriptsize$\pm$.036} & .274{\scriptsize$\pm$.008} & .280{\scriptsize$\pm$.023} & .271{\scriptsize$\pm$.003} & .047{\scriptsize$\pm$.012} & .123{\scriptsize$\pm$.023} & .172{\scriptsize$\pm$.023} & .188{\scriptsize$\pm$.023} \\
SDC & .435{\scriptsize$\pm$.023} & .436{\scriptsize$\pm$.010} & .456{\scriptsize$\pm$.008} & .451{\scriptsize$\pm$.018} & .446{\scriptsize$\pm$.015} & .444{\scriptsize$\pm$.010} & .441{\scriptsize$\pm$.012} & .430{\scriptsize$\pm$.002} & .276{\scriptsize$\pm$.013} & .321{\scriptsize$\pm$.007} & .319{\scriptsize$\pm$.026} & .317{\scriptsize$\pm$.006} & .171{\scriptsize$\pm$.090} & .285{\scriptsize$\pm$.013} & .292{\scriptsize$\pm$.011} & .290{\scriptsize$\pm$.010} \\
LDC & .421{\scriptsize$\pm$.012} & .425{\scriptsize$\pm$.017} & .457{\scriptsize$\pm$.007} & .451{\scriptsize$\pm$.017} & .435{\scriptsize$\pm$.020} & .441{\scriptsize$\pm$.009} & .442{\scriptsize$\pm$.008} & .426{\scriptsize$\pm$.006} & .270{\scriptsize$\pm$.016} & .290{\scriptsize$\pm$.006} & .294{\scriptsize$\pm$.032} & .294{\scriptsize$\pm$.005} & .057{\scriptsize$\pm$.004} & .099{\scriptsize$\pm$.040} & .127{\scriptsize$\pm$.020} & .134{\scriptsize$\pm$.015} \\
Frozen & .186{\scriptsize$\pm$.042} & .210{\scriptsize$\pm$.045} & .199{\scriptsize$\pm$.043} & .202{\scriptsize$\pm$.046} & .375{\scriptsize$\pm$.040} & .394{\scriptsize$\pm$.038} & .398{\scriptsize$\pm$.020} & .397{\scriptsize$\pm$.006} & .306{\scriptsize$\pm$.014} & .317{\scriptsize$\pm$.012} & .331{\scriptsize$\pm$.011} & .318{\scriptsize$\pm$.015} & .177{\scriptsize$\pm$.005} & .178{\scriptsize$\pm$.010} & .183{\scriptsize$\pm$.012} & .172{\scriptsize$\pm$.019} \\
\system{}-lite & .507{\scriptsize$\pm$.012} & .501{\scriptsize$\pm$.010} & .510{\scriptsize$\pm$.001} & \textbf{.512{\scriptsize$\pm$.020}} & \textbf{.470{\scriptsize$\pm$.004}} & \textbf{.481{\scriptsize$\pm$.014}} & \textbf{.473{\scriptsize$\pm$.008}} & \textbf{.454{\scriptsize$\pm$.008}} & .316{\scriptsize$\pm$.022} & .355{\scriptsize$\pm$.018} & \textbf{.356{\scriptsize$\pm$.019}} & \textbf{.356{\scriptsize$\pm$.014}} & .227{\scriptsize$\pm$.008} & .246{\scriptsize$\pm$.017} & .262{\scriptsize$\pm$.018} & .261{\scriptsize$\pm$.014} \\
\textbf{\system{}} & \textbf{.516{\scriptsize$\pm$.009}} & \textbf{.510{\scriptsize$\pm$.018}} & \textbf{.513{\scriptsize$\pm$.005}} & .509{\scriptsize$\pm$.017} & .463{\scriptsize$\pm$.001} & .470{\scriptsize$\pm$.011} & .464{\scriptsize$\pm$.005} & .448{\scriptsize$\pm$.011} & \textbf{.349{\scriptsize$\pm$.015}} & \textbf{.364{\scriptsize$\pm$.015}} & .355{\scriptsize$\pm$.019} & .355{\scriptsize$\pm$.015} & \textbf{.262{\scriptsize$\pm$.018}} & \textbf{.287{\scriptsize$\pm$.012}} & \textbf{.295{\scriptsize$\pm$.016}} & \textbf{.295{\scriptsize$\pm$.014}} \\
\cmidrule(l){1-17}
\textcolor{gray}{Oracle} & \textcolor{gray}{.505{\scriptsize$\pm$.009}} & \textcolor{gray}{.510{\scriptsize$\pm$.012}} & \textcolor{gray}{.509{\scriptsize$\pm$.005}} & \textcolor{gray}{.507{\scriptsize$\pm$.020}} & \textcolor{gray}{.423{\scriptsize$\pm$.011}} & \textcolor{gray}{.430{\scriptsize$\pm$.014}} & \textcolor{gray}{.432{\scriptsize$\pm$.011}} & \textcolor{gray}{.418{\scriptsize$\pm$.008}} & \textcolor{gray}{.330{\scriptsize$\pm$.018}} & \textcolor{gray}{.342{\scriptsize$\pm$.014}} & \textcolor{gray}{.333{\scriptsize$\pm$.021}} & \textcolor{gray}{.336{\scriptsize$\pm$.020}} & \textcolor{gray}{.298{\scriptsize$\pm$.005}} & \textcolor{gray}{.287{\scriptsize$\pm$.008}} & \textcolor{gray}{.297{\scriptsize$\pm$.014}} & \textcolor{gray}{.282{\scriptsize$\pm$.006}} \\
\bottomrule
\end{tabular}}
\caption{Rare-class F1 for different backbones on nuImages dataset ($R{=}100$).}
\label{tab:backbone_grid}
 \vspace{-3ex}
\end{table*}
%%%%%%%%%%%%%%%%%%%%%%%%%%%%%%%

\subsection{Generalization under Various Configurations}
\label{sec:eval_generalizaiton}

%%%%%%%%%%%%%%%%%%%%%%%%%%%%%%%

To assess \system{}'s performance in general and diverse settings, we compare with five representative baselines across three classification tasks, four backbone architectures, varying remaining rare-class sample fractions from 1\% to 10\%. 

Table~\ref{tab:dataset_var} reports rare-class F1 scores for nuImages, coarse-grained Car-1000, and a more challenging fine-grained Car-1000 task using the three-layer CNN. For the fine-grained task, we use samples belonging to the sedan category and classify its subcategories by vehicle size. As described earlier, these tasks progress from diverse road objects to vehicle types that are visually similar. Overall, \system{} variants achieve the highest F1 in all configurations. \system{} leads in 11 of the 12 settings, while \system{}-lite is numerically higher on nuImages at 10\% remaining rare-class samples by only 0.3\%p, well within the reported variation. \system{} shows its clearest margin over the strongest non-\system{} baseline on fine-grained Car-1000 (4.9\%p at 1\%, compared with 8.1\%p on nuImages and 3.2\%p on coarse-grained Car-1000). This pattern is even clearer relative to \system{}-lite, which remains below the strongest non-\system{} baseline across all remaining-sample fractions on the fine-grained task, while \system{} consistently performs the best. This suggests that covariance-based residual adaptation becomes valuable when the rare class is similar to remaining classes and the scalar correction of \system{}-lite is insufficient. Note that a few results outperform the oracle. This is possible because the oracle uses the current mean of the test data rare-class features and is therefore a reference for prototype alignment, not a guaranteed upper bound on F1. When classes overlap, a slightly drifted prototype may produce a better balance between precision and recall.

Table~\ref{tab:backbone_grid} reports the architecture-level results on nuImages using three-layer CNN, ResNet-18, MobileNetV3, and ViT-Tiny, covering both CNN- and transformer-based backbones practical for in-vehicle training. ViT-Tiny follows the original ViT design and uses the final class token as the feature representation rather than patch-token representation pooling~\cite{dosovitskiy2020image}. \system{} variants achieve the highest F1 in all configurations, showing that its benefit is not tied to a particular architecture.

A notable observation is that \system{}-lite remains competitive with \system{} across CNN backbones, whereas the performance gap becomes substantially larger on ViT-Tiny. We conjecture that this difference arises from the learned feature space geometry. CNN representations have been reported to be relatively linear~\cite{kim2024curved}, while ViTs form more curved representation spaces due to nonlinear attention operations. Under such geometry, representation drift is less likely to be captured by the common scale adjustment used in \system{}-lite, emphasizing the importance of directional information provided by the covariance-based transport in \system{}. Thus, \system{}-lite remains an effective lightweight option for backbones with near-linear feature drift, but its suitability depends on the underlying model's representation geometry. Overall, \system{} remains effective across task difficulty, backbone family, and rare-class availability, with \system{} providing the most consistent performance and \system{}-lite retaining competitive results at lower communication cost.

\subsection{Method Analysis}

In this section, we analyze whether each component of \system{} reduces the prototype reconstruction error as it was designed to address. Unlike our preliminary studies (Sec.~\ref{sec:prelim}), which established the occurrence of post-departure drift, this analysis isolates how in-span tracking and out-of-span correction contribute to the final reconstruction.

\begin{figure}[t]
    \centering
    \includegraphics[width=\columnwidth]{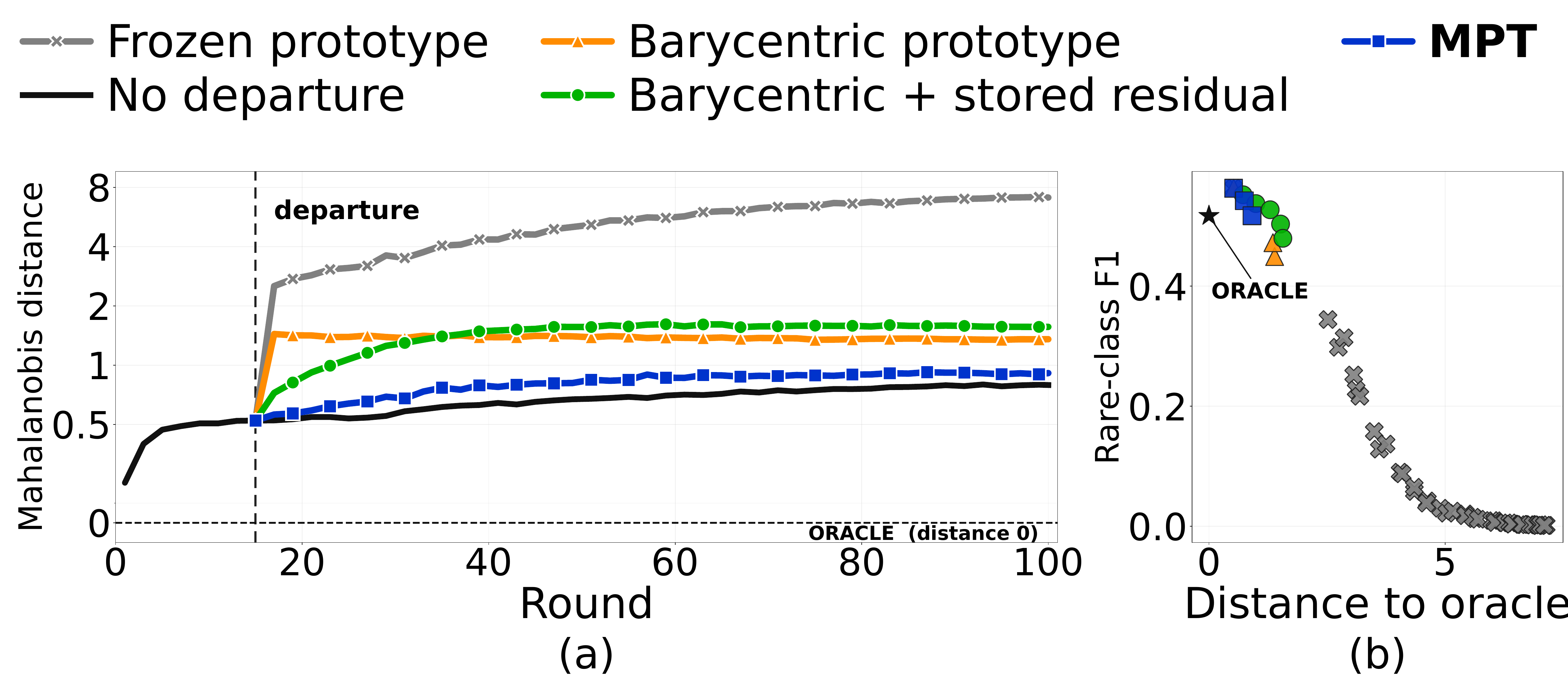}
    \caption{%
    Prototype reconstruction error (nuImages) and rare-class F1 correlation.
    (a) Mahalanobis distance of prototypes to oracle.
    (b) Rare-class F1 against prototype distance to oracle.
    }
    \label{fig:proto_quality}
    \vspace{-2ex}
\end{figure}

To characterize how prototype reconstruction evolves as each component of \system{} is added, Figure~\ref{fig:proto_quality}~(a) compares four reconstruction settings in sequence. ``Frozen Prototype'' retains the prototype observed at departure, ``Barycentric Prototype'' reconstructs only the in-span component from the updated remaining-class prototypes, and ``Barycentric + Stored Residual'' further adds the out-of-span residual stored at departure. \system{} additionally updates this residual as training continues. The plots report the Mahalanobis distance from each reconstruction to the current oracle prototype, where a smaller distance indicates better alignment. Figure~\ref{fig:proto_quality}~(b) relates this distance to rare-class F1 and shows that prototypes closer to the oracle achieve higher F1. Therefore, we use the distance as the primary measure of reconstruction error in the following analysis.
In Figure~\ref{fig:proto_quality}~(a), the gray ``Frozen Prototype'' plot moves steadily farther from the oracle as continued backbone updates change the feature space. The orange ``Barycentric Prototype'' plot remains relatively stable, indicating that the barycentric relationship tracks the in-span drift shared with the remaining classes. However, there is a clear non-closing gap due to the lack of the out-of-span component. The green ``Barycentric + Stored Residual'' plot adds this missing component and substantially reduces the initial gap. Its error rises again in later rounds as the fixed departure-time residual becomes outdated. These results show that barycentric reconstruction tracks in-span drift, while sustained alignment requires out-of-span residual updates.

% Figure~\ref{fig:proto_quality}
\begin{figure}[t]
    \centering
    \includegraphics[width=\columnwidth]{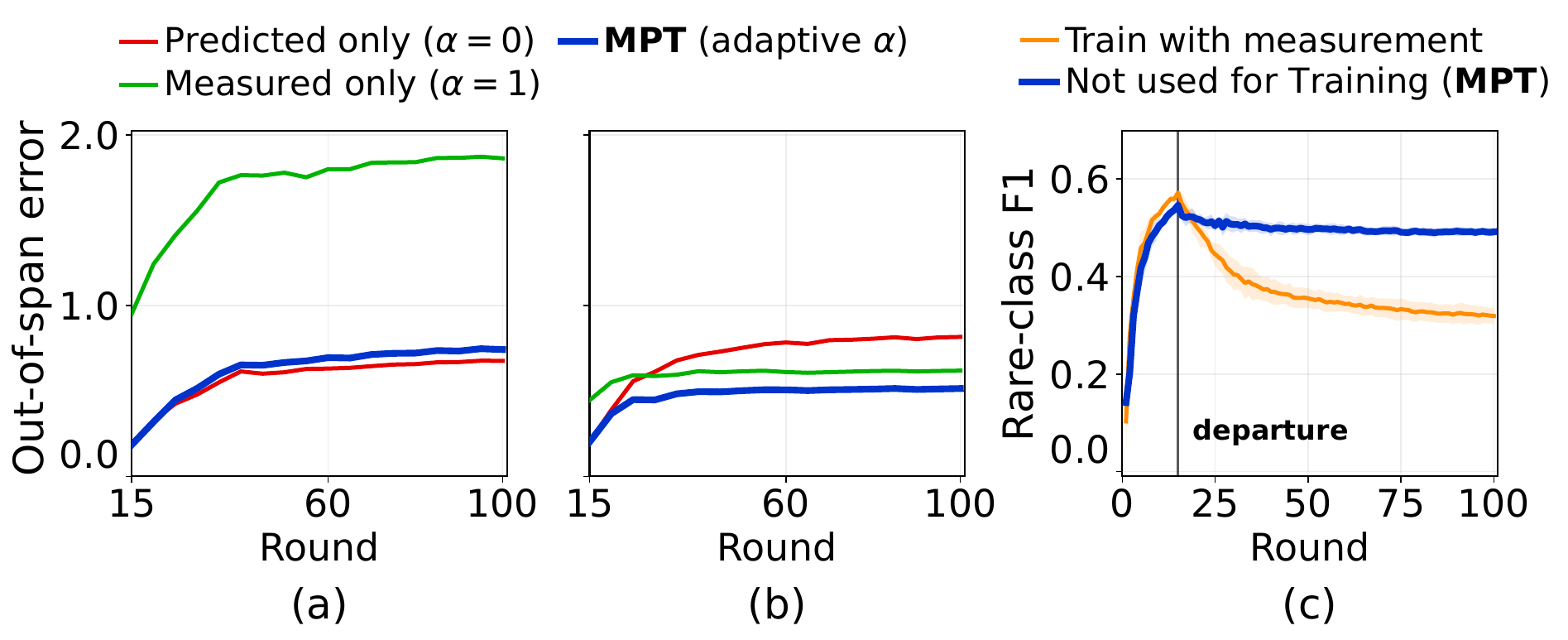}
    \caption{Out-of-span reconstruction error with different rare-class sample budgets. (a) and (b) 2\% and 20\% of the rare-class samples remaining, respectively (nuImages; three-layer CNN). (c) Per round F1 difference on how remaining rare data used.}
    \label{fig:out-of-span-eval}
    \vspace{-2ex}
\end{figure}

% \noindent\textbf{Correcting Out-of-Span Drift.}
Having shown that the stored residual becomes outdated, we next analyze how \system{} updates the missing out-of-span component and how effectively its two correction sources reduce the resulting error. Figures~\ref{fig:out-of-span-eval}~(a) and (b) report the distance between each reconstructed out-of-span residual and the current oracle residual, where a smaller value indicates more accurate reconstruction. The red ``Predicted Only'' plot disables measurement with \(\alpha=0\) and updates the residual only from covariance changes. The green ``Measured Only'' plot disables prediction with \(\alpha=1\) and uses only the residual measured from the remaining rare-class samples. In practice, \system{} (blue), adjusts \(\alpha\) to combine the two sources according to their reliability. With only 2\% of rare-class samples remaining in Figure~\ref{fig:out-of-span-eval}~(a), the measured residual is highly noisy due to low sample quantity. Consequently, \system{} places greater weight on prediction and stays close to its lower error. When 20\% remain in Figure~\ref{fig:out-of-span-eval}~(b), measurement becomes more reliable, and combining them with prediction leads to lower error than either source alone for \system{}. These results show that adaptive weighting allows \system{} to use the more reliable source under each sample budget, making it effective across different levels of rare-class data availability.

% \noindent\textbf{Small Rare Data Usage.}
Finally, Figure~\ref{fig:out-of-span-eval}~(c) examines why \system{} reserves the remaining rare-class samples for prototype measurement rather than reusing them for continued backbone training. The figure compares \system{}, which keeps these samples out of training, with a variant that uses the same samples for both training and measurement. Both perform similarly near departure, but the training variant gradually loses rare-class F1 as FL continues. Under the highly skewed post-departure distribution, remaining rare-class samples contribute only limited updates compared with the remaining classes. Reusing them for training also makes their features less suitable as an independent reference for measuring rare-class-specific drift. These results support the design choice of excluding the remaining rare-class samples from backbone training and using them only to calibrate the reconstructed prototype.

\section{Discussions}\label{sec:discussion}

\noindent\textbf{Task generalization.}
Our evaluation focuses on maintaining image classification performance after a vehicle holding most samples of a rare class departing from the FL network. However, the same post-departure (or client drop) degradation can arise in other perception tasks as well. In general object detection tasks, the ``forgetting'' behavior of a neural network has been shown to affect the region classifier more strongly than the localization branch~\cite{wu2025demystifying}, which agrees with our observations in Section~\ref{sec:prelim}. To support object detection tasks, \system{} could benefit by replacing the classification branch of the detection head, which can help preserve recognition of the rare class after departure. We see the extension of our work to such tasks as a meaningful research direction. %\system{} could also support more complex participation changes, a joining or rejoining vehicle would resume direct prototype updates using its class-level statistics, while multiple affected classes could be handled through separate prototype reconstructions.
%
% In this case, \system{} would replace the region classifier with the reconstructed prototype.%Extending \system{} to detection would require adapting the reconstructed prototype to a trainable detection head.
% The framework can also support more complex participation changes, joining or rejoining vehicles can resume direct prototype updates using its class-level statistics, while multiple affected classes can be handled through separate prototype reconstructions.

\vspace{0.5ex}
\noindent\textbf{Obtaining statistics before departure.}
\system{} requires exchanging current class-level statistics of a vehicle before it leaves the FL network. These are already reported each round, since each vehicle re-embeds its local data and returns class-wise feature statistics and sample counts.
%\system{} requires current class-level statistics before a vehicle leaves the FL cluster, and these statistics are already computed and reported as part of the workflow described in Section~\ref{sec:overall}, where each vehicle re-embeds its local data and reports class-wise feature statistics and sample counts.
This requirement is generally feasible in charging scenarios, where a local departure event, such as a vehicle-start event, 
triggers the vehicle to upload its final statistics before leaving the FL network.
%
%This requirement is feasible in the vehicle charging scenario, where a local departure event such as engine-start activities can trigger a final statistics upload before disconnection. 
%
The server can then determine whether the departing vehicle holds most samples of a class. If its departure leaves only a few samples of that class among the remaining vehicles, the server stores the barycentric weights, residual, and covariance reference required for subsequent prototype reconstruction.

%\jk{Can we come up with something that links to future research directions for others?}

\vspace{0.5ex}
\noindent\textbf{Toward more realistic mobility dynamics.}
This work mainly focuses on a controlled parking-lot scenario to isolate and clearly evaluate the impact of permanent vehicle departure. In real deployments, however, vehicles may arrive and depart dynamically, and their local data distributions may vary substantially with respect to their routes and driving conditions. A more realistic evaluation could model vehicle arrivals and departures using stochastic processes, such as a Poisson process, and derive client-specific label distributions from data collected along real driving routes. We leave the evaluation of \system{} under such realistic mobility and data-distribution dynamics to future work.

\section{Conclusion}\label{sec:conclusion}

This work proposes \system{}, a lightweight framework for preserving rare-class recognition in cross-vehicle FL under heterogeneous data distributions and dynamic vehicle participation. \system{} combines prototype-based NCM with post-departure prototype reconstruction that tracks the in-span component through a barycentric relationship and adapts the out-of-span residual using covariance changes and remaining rare-class statistics. Across three vehicle-classification tasks, four backbones, and also varying remaining-sample conditions, \system{} consistently outperforms five representative baselines while closely tracking the oracle prototype. Our results show that \system{} preserves rare-class recognition after departure under FL privacy constraints, without extra retraining.

% [double-blind] Acknowledgment는 camera-ready에서만 (자금/이름 식별 금지):
% \section*{Acknowledgment}
% This work was supported by \ldots

\section{Use of AI Disclosure}
% We used Anthropic Claude to assist with manuscript editing and experimental code development. Its outputs did not determine the study design or conclusions. AI-assisted content and code were reviewed, tested, and verified by the authors.

We used Anthropic Claude to assist with manuscript editing and experimental code development. AI outputs were used only as suggestions under supervision and did not determine the study design or conclusions. All AI-assisted content and code were reviewed, tested, and verified by the authors.

%---------------- 참고문헌 ----------------
\bibliographystyle{IEEEtran}
\bibliography{reference}
\end{document}